\documentclass[runningheads]{llncs}
\usepackage[T1]{fontenc}
\usepackage{graphicx}
\usepackage{hyperref}
\usepackage{color}

\usepackage{tikz}
\usetikzlibrary{shapes, arrows, positioning}
\usepackage{graphicx}
\usepackage{subcaption}

\usepackage{comment}

\usepackage{algorithm}
\usepackage{algpseudocode}

\usepackage{amsmath}
\usepackage{amsfonts}

\usepackage{bm}

\begin{document}
\title{Active Inference Meeting Energy-Efficient Control of Parallel and Identical Machines\thanks{{Accepted at the 10th International Conference on Machine Learning, Optimization, and Data Science.}}}
\titlerunning{AIF Meeting EEC}

\author{Yavar Taheri Yeganeh\inst{1} \and
Mohsen Jafari\inst{2} \and
Andrea Matta\inst{1}}

\institute{1. Politecnico di Milano, Milan, Italy \\
\email{\{yavar.taheri, andrea.matta\}@polimi.it} \and
2. Rutgers University, Piscataway, USA \\
\email{jafari@soe.rutgers.edu}}

\authorrunning{Y. Taheri Yeganeh et al.}

\maketitle              
\begin{abstract}
We investigate the application of active inference in developing energy-efficient control agents for manufacturing systems. Active inference, rooted in neuroscience, provides a unified probabilistic framework integrating perception, learning, and action, with inherent uncertainty quantification elements. Our study explores deep active inference, an emerging field that combines deep learning with the active inference decision-making framework. Leveraging a deep active inference agent, we focus on controlling parallel and identical machine workstations to enhance energy efficiency. We address challenges posed by the problem's stochastic nature and delayed policy response by introducing tailored enhancements to existing agent architectures. Specifically, we introduce multi-step transition and hybrid horizon methods to mitigate the need for complex planning. Our experimental results demonstrate the effectiveness of these enhancements and highlight the potential of the active inference-based approach.

\keywords{Active Inference \and Probabilistic Deep Learning \and Reinforcement Learning \and Energy-Efficient Control \and Manufacturing Systems}
\end{abstract}
\section{Introduction}
\label{sec:introduction}
Active inference (AIF), an emerging field inspired by the principles of biological brains, offers a promising alternative for decision-making models. It unifies perception, learning, and decision-making under the free energy principle (FEP), which formulates neuronal inference and learning under uncertainty \cite{friston2010free}. Accordingly, the brain is modeled through levels of (variational) Bayesian inference \cite{parr2022active}, minimizing prediction errors by leveraging a generative model of the world while considering uncertainties. This framework enables the development of agents that can calibrate their models and make decisions without complete knowledge of system dynamics. Significant progress has been made in applying active inference across various domains, including robotics, autonomous driving, and healthcare \cite{pezzato2023active,schneider2022active,huang2024navigating}, showcasing its ability to handle complex decision-making tasks in dynamic environments. 
Recently, the manufacturing industry's focus on energy efficiency has intensified due to its significant contribution to global energy consumption. Addressing energy efficiency at the machine level has become critical, with energy-efficient scheduling (EES) and energy-efficient control (EEC) strategies emerging as key approaches to reducing environmental impact \cite{loffredo2024energy}. While traditional EEC methods often necessitate complete system knowledge, reinforcement learning (RL) has shown potential in optimizing manufacturing processes without prior system knowledge \cite{LOFFREDO202391,loffredo2023reinforcement}. However, RL agents may struggle to rapidly adjust their policies to changing conditions. 

This research aims to build upon advancements in active inference-based decision-making \cite{Fountas2020DeepAI,da2023reward} and apply it to EEC in manufacturing systems, demonstrating its potential and advancing the understanding of active inference in complex environments. The existing active inference agents often rely on extensive search algorithms during planning and make decisions based on immediate next predictions \cite{Fountas2020DeepAI}, which pose challenges in the context of the EEC problem. By employing deep active inference as the decision-making algorithm, we introduce tailored enhancements, such as multi-step transition and hybrid horizon methods, to address the challenges posed by the problem's stochastic nature and delayed policy response. Our experimental results highlight the effectiveness of these enhancements and underscore the potential of the active inference-based approach. The remainder of the paper is organized as follows: we begin by concisely introducing the EEC problem and the manufacturing system under study. We then present an overview of the formalization of active inference, describe the agent, evaluate its performance, and discuss future directions.

\section{Application Overview}
\label{Application}
Active inference has proven effective in various applications commonly associated with decision-making processes in biological agents, such as humans and animals. These applications primarily involve visual sensory output as observations. For instance, Fountas et al. (2020) \cite{Fountas2020DeepAI} tested their agent on tasks like \textit{Dynamic dSprites}\cite{higgins2016beta} and \textit{Animal-AI}\cite{crosby2019animal}, which can be performed by biological agents simply. Additionally, applications in robotics \cite{lanillos2021active,da2022active} (e.g., manipulation \cite{schneider2022active}) align with tasks that human agents can typically perform naturally. This effectiveness stems from active inference being a theory of decision-making for biological agents \cite{parr2022active}. However, certain applications, such as the control of industrial systems, can present complex challenges. While the decision-making processes and existing applications mentioned above may not be straightforward, human agents may struggle to devise effective policies for these more intricate problems.

\subsubsection{EEC in Manufacturing Systems}
\label{subsec:eec}
EEC is gaining prominence in both academic and industrial circles within manufacturing systems. It offers substantial energy savings across three key control levels: component, machine, and production system. At its core, EEC involves managing the power consumption state of objects based on environmental conditions. Objects are kept fully operational when their functions are needed and transitioned to low power states when not in use, though this poses challenges due to the unpredictable nature of demands and the penalties incurred during state transitions. These penalties include both time wasted during transition, when the object adds no value, and the energy consumed during the transition process. A comprehensive and recent literature review on this topic can be found in \cite{renna2021literature}.

\subsubsection{System Description}
\begin{figure}[!h]
    \centering
    \includegraphics[width=\textwidth]{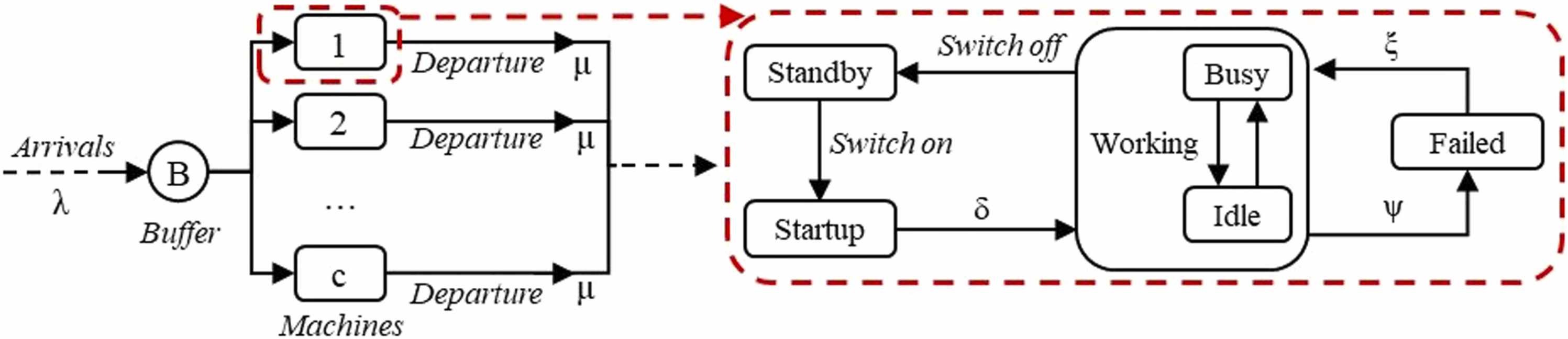}
    \caption{Layout of parallel and identical machines in the workstation \cite{LOFFREDO202391}.}
    \label{fig:eec}
\end{figure}

We follow the task outlined by Loffredo et al. (2023) \cite{LOFFREDO202391}, which focuses on a stand-alone manufacturing workstation that can be extended to more complex multi-stage production lines \cite{loffredo2024energy}. Accordingly, the system under study has an upstream buffer \(B\) with finite holding capacity serving multiple identical parallel machines, as depicted in Fig. \ref{fig:eec}. It is subject to the stochastic arrival of parts, while machines can transition between various states: working (idle or busy), standby, startup, and failed. Power consumption varies by state: standby (\(w_{sb}\)), failed (\(w_{f}\)), startup (\(w_{su}\)), idle (\(w_{id}\)), and busy (\(w_{b}\)), where \(w_b > w_{su} > w_{id} > w_{sb} \approx w_f \approx 0\).
The key characteristic of this system is that all processes are stochastic, modeled as Poisson processes \cite{kingman1992poisson}. This pertains to the arrival rate (\(\lambda\)) to buffer \(B\), machine processing times (\(\mu\)), startup times (\(\delta\)), time between failures (\(\psi\)), and time to repair (\(\xi\)), all with expected values, independent and stationary. Machines work on a single part type following a first-come-first-served rule and cannot be switched off during processing or startup. Machines become starved (i.e., enter an idle state) if they are ready to process but \(B\) is empty, and they cannot be blocked as there is an infinite downstream buffer. Effective EEC aims to manage power variations efficiently to minimize energy consumption without compromising productivity. Therefore, the primary task is to dynamically decide the number of machines to keep working or \emph{switch off} based on the emergent stochastic patterns, ensuring optimal trade-offs between energy savings and system throughput.

\section{Deep Active Inference Agent}
\label{sec:daif}
Active inference is a unifying theory that integrates inference, perception, and action by emphasizing the dependence of observations on actions \cite{millidge2022predictive}. To effectively manage observations toward preferred states, the optimization of actions plays a crucial role \cite{millidge2022predictive}. This concept was originally proposed as a framework for understanding how organisms actively control and navigate their environment by iteratively updating their beliefs and actions based on sensory evidence \cite{parr2022active}. The FEP \cite{friston2010action,millidge2021applications} is at the core of active inference, paving the way for creating a mathematical model, and there are even experimental evidences supporting it \cite{isomura2023experimental}. We modify the proposed agent by Fountas et al. (2020) \cite{Fountas2020DeepAI}, which exhibited notable capabilities and performance when compared against three benchmark model-free RL algorithms in two image applications.

\subsection{Active Inference Formalism}
\label{subsec:aif}

Active inference agents employ an integrated probabilistic framework consisting of an internal generative model \cite{da2023reward} coupled with inference mechanisms to represent and interact with the world. Similar to RL the agent interacts with the environment, but using three random variables representing observation, latent state, and action (i.e., $(o_{t},s_{t},a_{t})$ at time $t$). The framework assumes a Partially Observable Markov Decision Process (POMDP) \cite{kaelbling1998planning,da2023reward,paul2023efficient}.
The generative model of the agent, parameterized with $\theta$, is defined over these variables (i.e., \(P_{\theta}(o_{1:t},s_{1:t},a_{1:t-1})\)) \cite{Fountas2020DeepAI}. Generally, the agent acts to reduce \textit{surprise}, which can be quantified by \(-\log P_{\theta}(o_{t})\). 
Specifically, there are two steps for the agent while interacting with the world \cite{parr2022active,Fountas2020DeepAI} as follows:

1) The agent calibrates its generative model through fitting predictions and improve its representation of the world. This is done by minimizing Variational Free Energy (VFE), which is similar to \textit{surprise} of predictions in connection with the actual observations \cite{Fountas2020DeepAI,sajid2022bayesian,paul2023efficient}, as follows:
\begin{equation}
\label{eqn:1}
{\theta}^{*} = \arg\min_{\substack{\theta}} \left(\mathbb{E}_{Q_{\phi}(s_{t},a_{t})}\left[\log Q_{\phi}(s_t,a_{t})-\log P_{\theta}(o_{t},s_{t},a_{t})\right]\right) \ .
\end{equation}
This objective function is commonly known as the negative evidence lower bound (ELBO) \cite{blei2017variational}, which is the upper bound for $-\log P_{\theta}(o_{t})$. It is also used as a foundation for training variational autoencoders \cite{kingma2013auto}.

2) The agent makes decisions (i.e., chooses actions) in active inference based on the accumulated negative Expected Free Energy (EFE or $G$):
\begin{equation}
\label{eqn:2}
P(\pi)\stackrel{} {=} {\sigma}( -G(\pi)) \ {=} \ \sigma \left(-\sum_{\tau>t}^{}{ G}(\pi,\tau)\right) \ ,
\end{equation}
where $\sigma(\cdot)$ represents the \emph{Softmax} function, and $\pi$ (i.e., policy) denotes the sequence of actions. The EFE encompasses minimizing \textit{surprise} regarding preferred observations \footnote{In EFE, the \textit{surprise} of predictions is measured with respect to the preference, while in VFE, it is measured with respect to the actual observation used to calibrate the model.},exploring uncertainty, and reducing uncertainty about model parameters \cite{Fountas2020DeepAI}. The EFE for $\tau \geq t$ can be formulated as follows \cite{schwartenbeck2019computational}:
\begin{equation}
\label{eqn:3}
G(\pi,\tau)=\mathbb{E}_{{ P}(\sigma_{\tau}|s_{\tau},\theta)}\mathbb{E}_{{\cal Q}_{\phi}(s_{\tau},\theta|\pi)}\left[\log Q_{\phi}(s_{\tau},\theta|\pi)-\log{ P}(o_{\tau},s_{\tau},\theta|\pi)\right] \ .
\end{equation}
Fountas et al., (2020) \cite{Fountas2020DeepAI} provided a derivation \cite{schwartenbeck2019computational} for calculating the EFE in Eq. \ref{eqn:3} at each time step:
\begin{subequations}
\label{eqn:efe-base}
\begin{align}
G(\pi,\tau) &= -\mathbb{E}_{\tilde{Q}}\left[\log P(o_{\tau}|\pi)\right] \label{eqn:efe-base-a} \\
&\quad + \mathbb{E}_{\tilde{Q}}\left[\log Q(s_{\tau}|\pi) - \log P(s_{\tau}|o_{\tau},\pi)\right] \label{eqn:efe-base-b} \\
&\quad + \mathbb{E}_{\tilde{Q}}\left[\log Q(\theta|s_{\tau},\pi) - \log P(\theta|s_{\tau},o_{\tau},\pi)\right] \ . \label{eqn:efe-base-c} 
\end{align}
\end{subequations}
They expanded the formalism, leading to a tractable estimate for EFE that is both interpretable and calculable \cite{Fountas2020DeepAI}:
\begin{subequations}
\label{eqn:efe}
\begin{align}
G(\pi,\tau) &= -\mathbb{E}_{Q(\theta|\pi)Q(s_{\tau}|\theta,\pi)Q(o_{\tau}|s_{\tau},\theta,\pi)}\left[\log\textstyle P(o_{\tau}|\pi)\right] \label{eqn:efe-a} \\
&\quad + \mathbb{E}_{Q(\theta|\pi)}\left[\,\mathbb{E}_{Q(o_{\tau} | \theta,\pi)}H(s_{\tau}|o_{\tau},\pi)-H(s_{\tau}|\pi)\right] \label{eqn:efe-b} \\
&\quad + \mathbb{E}_{Q(\theta|\pi)Q(s_{\tau}|\theta,\pi)}H(o_{\tau}|s_{\tau},\theta,\pi) - \mathbb{E}_{Q(s_{\tau}|\pi)}H(o_{\tau}|s_{\tau},\pi) \ . \label{eqn:efe-c}
\end{align}
\end{subequations}
This paved the way for establishing a unified formalism for computing decisions in Eq. \ref{eqn:2}. Accordingly, actions are connected to perception, which is achieved through Bayesian inference, based on the EFE in Eq. \ref{eqn:efe}.

Through this formalism that leads to calculating the EFE (i.e., Eq. \ref{eqn:efe}), we can interpret the contribution of each element \cite{Fountas2020DeepAI}: \textbf{The first term} (i.e., Eq. \ref{eqn:efe-a}) is analogous to reward in RL as it is the \textit{surprise}\footnote{Here, instead of maximizing cumulative rewards, the focus is on minimizing the \textit{surprise}, which quantifies the extent of deviation (i.e., misalignment) between the prediction and the preferred observation.} of the prediction considering preferred observation. \textbf{The second term} (i.e., Eq. \ref{eqn:efe-b}) represents state uncertainty, which is mutual information between the agent’s beliefs about state before and after prediction. This term also shows a motivation to explore areas of the environment that resolve state uncertainty \cite{Fountas2020DeepAI}. \textbf{The third term} (i.e., Eq. \ref{eqn:efe-c}) represents uncertainty about model parameters considering new observations. This term is also referred active learning, novelty, or curiosity \cite{Fountas2020DeepAI}. In fact, model parameters (i.e., $\theta$), particularly contribute to making predictions, including generation of the next states. 

\begin{figure}[H]
\centering
\begin{subfigure}{0.49\textwidth}
  \includegraphics[width=\textwidth]{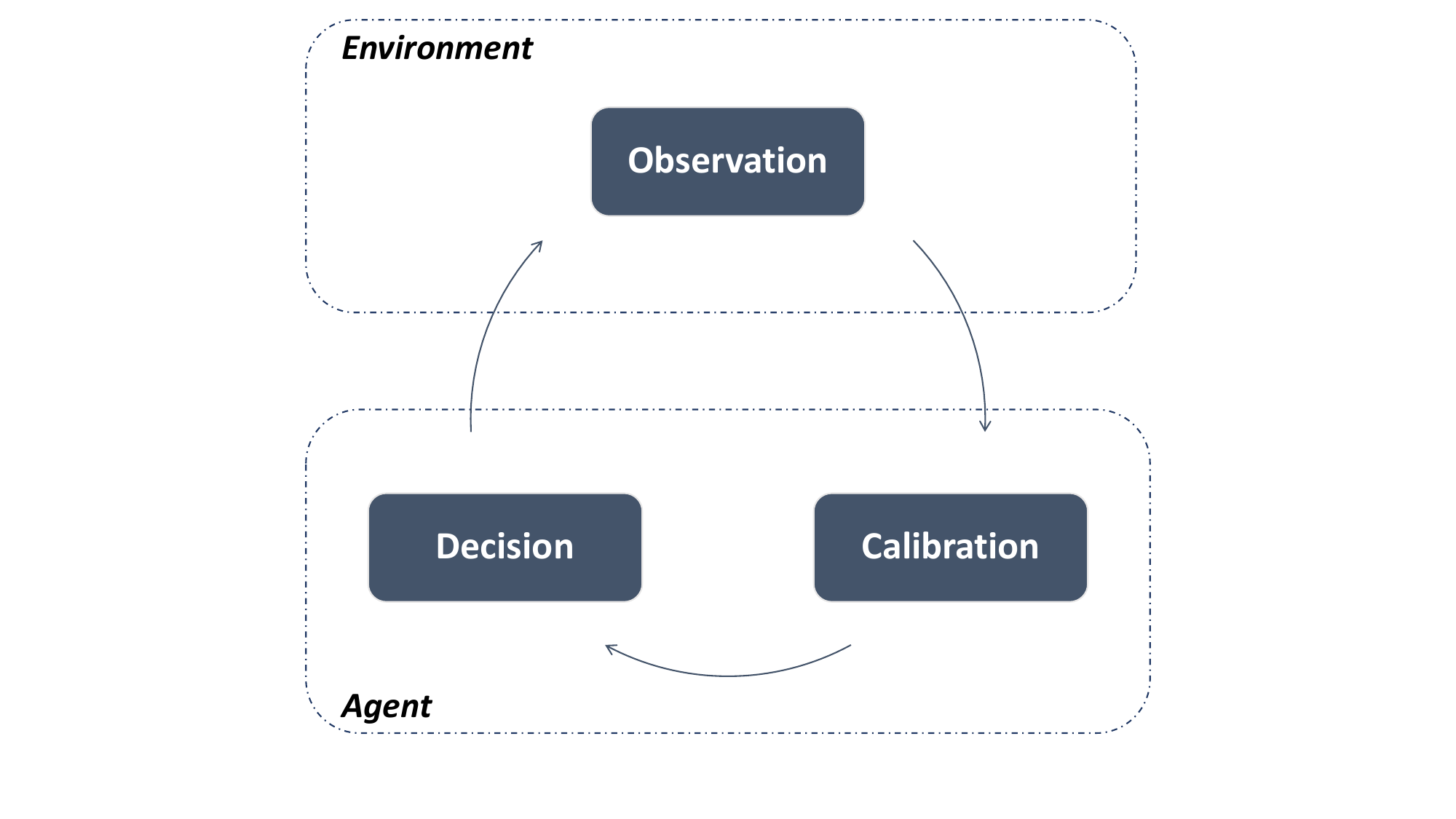}
\end{subfigure}
\begin{subfigure}{0.49\textwidth}
  \includegraphics[width=\textwidth]{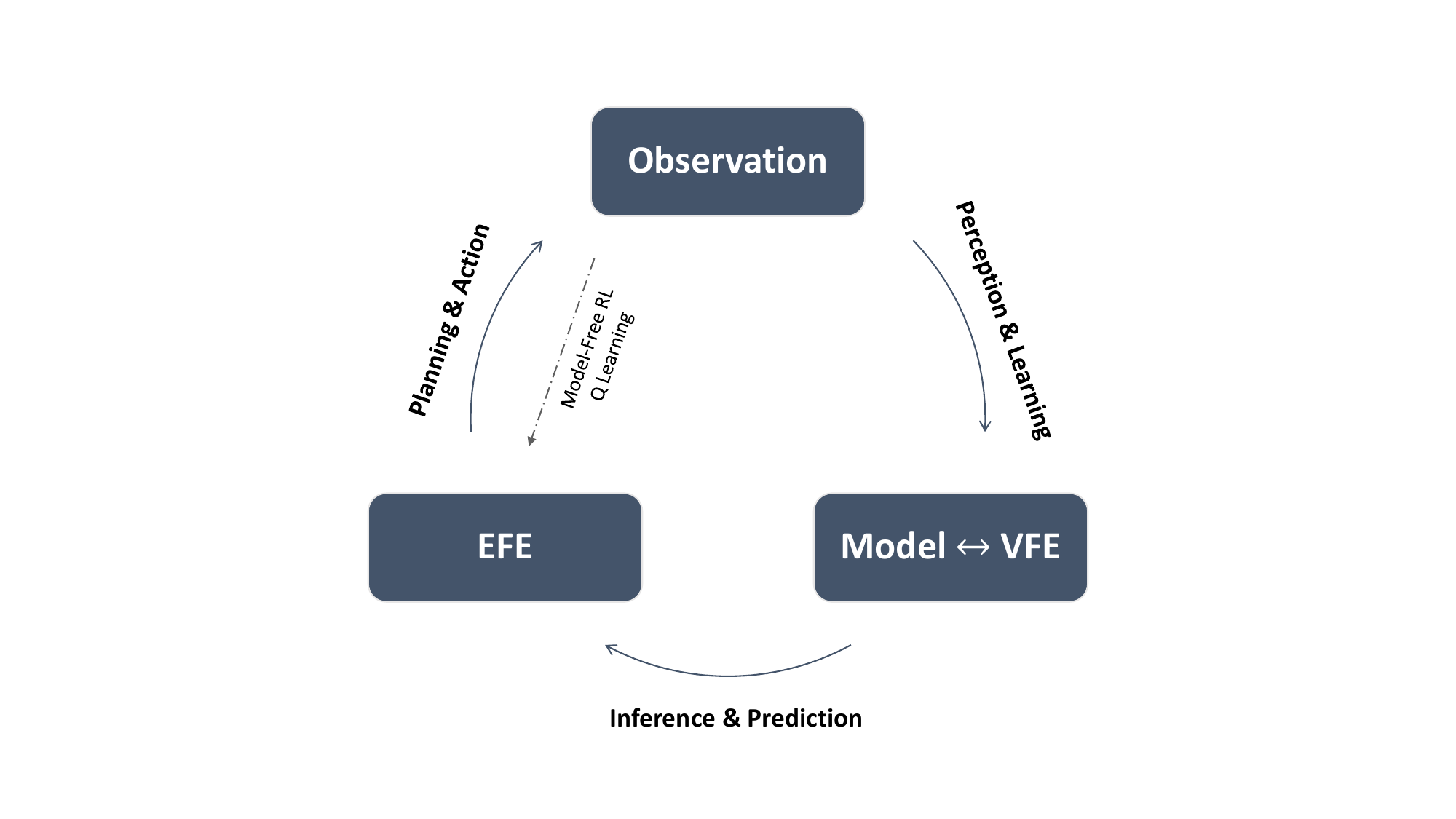}
\end{subfigure}
\caption{The illustration depicts two views of the active inference framework: general steps on the left and active inference elements on the right.}
\label{fig:framework}
\end{figure}

In summary, the framework (as depicted in Fig. \ref{fig:framework}) is realized through a mathematical formalism  in the following manner. The observation is fed as input, which propagates through the model to create perception (i.e., beliefs), which include generating future states. It is to facilitate the calculation of EFE (in Eq. \ref{eqn:efe}) integrated into the planner derived from policy (in Eq. \ref{eqn:2}) to act on the environment. After obtaining the next observation the VFE (in Eq. \ref{eqn:1}) can be calculated, which calibrate (i.e., learning) the model based on the matching the new observation with the prediction. In fact, every time the framework goes through the loop the model is optimized based on the VFE form the previous loop, then the optimized model is used to for the rest.

\subsection{Architecture}
\label{subsec:architecture}
An agent within the active inference framework needs different modules, which are entangled within the framework. Amortization \cite{kingma2013auto,marino2018iterative,gershman2014amortized} is introduced into the formalism (in Sec. \ref{subsec:aif}) to scale-up the realization \cite{Fountas2020DeepAI}. The formalism is inherently probabilistic and parameterized with with two sets, $\theta = \{{\theta}_{s}, {\theta}_{o}\}$ representing generative and $\phi = \{{\phi}_{s}\}$ recognition elements \cite{Fountas2020DeepAI}. Using the following parameterized modules the formalism (in Sec. \ref{subsec:aif}) can be calculated: \textbf{Encoder} (i.e., $Q_{\phi_{s}}(s_{t})$), an amortized inference of the hidden state (i.e., an inference network \cite{margossian2023amortized} providing a mapping between the observation, $\tilde{o}_{t}$, and a distribution for its corresponding hidden state). \textbf{Transition} (i.e., ${P}_{\theta_{s}}({s}_{t+1}|\tilde{s}_{t},\tilde{a}_{t})$), which  generates a distribution for the next hidden state based on both a sampled action and the current hidden state. \textbf{Decoder} (i.e., ${P}_{\theta_{o}}({o}_{t+1}|\tilde{s}_{t+1})$) generates a distribution for prediction based on the sampled hidden state. 

Neural networks can facilitate the realization of these modules by representing a mapping between a sample and the respective distribution. In fact, parameters of a pre-selcted (e.g., Gaussian) distribution can be approximated. Here, we model the state space specifically with a multivariate Gaussian distribution, assuming no covariance (i.e., diagonal Gaussian). Using the VFE (i.e., Eq. \ref{eqn:1}), all the three networks can be trained in an end to end fashion. Aside from action and the transition, which can even be considered integrated within the state space, the structure bears a resemblance to a variational autoencoder \cite{kingma2013auto}. It is noteworthy that training for the both include optimizing ELBO, as specified in Eq. \ref{eqn:1}.

\begin{figure}[!htbp]
    \centering
    \begin{tikzpicture}[x=0.75pt,y=0.75pt,yscale=-1,xscale=1]
    
    \draw  [color={rgb, 255:red, 0; green, 0; blue, 0 }  ,draw opacity=1 ][fill={rgb, 255:red, 74; green, 74; blue, 74 }  ,fill opacity=1 ] (477.71,207.54) -- (409.6,191.99) -- (408.11,71.99) -- (475.81,53.82) -- cycle ;
    \draw  [fill={rgb, 255:red, 74; green, 74; blue, 74 }  ,fill opacity=1 ] (248.44,71.82) -- (353.82,71.82) -- (353.82,188.05) -- (248.44,188.05) -- cycle ;
    \draw  [fill={rgb, 255:red, 74; green, 74; blue, 74 }  ,fill opacity=1 ] (128.47,54.56) -- (196.43,71.24) -- (196.64,191.25) -- (128.74,208.29) -- cycle ;
    \draw  [fill={rgb, 255:red, 155; green, 155; blue, 155 }  ,fill opacity=1 ] (213.19,71.82) -- (230.82,71.82) -- (230.82,188.05) -- (213.19,188.05) -- cycle ;
    \draw  [fill={rgb, 255:red, 155; green, 155; blue, 155 }  ,fill opacity=1 ] (371.82,71.82) -- (387.82,71.82) -- (387.82,188.05) -- (371.82,188.05) -- cycle ;
    \draw  [fill={rgb, 255:red, 155; green, 155; blue, 155 }  ,fill opacity=1 ] (89.82,56.22) -- (107.44,56.22) -- (107.44,209.82) -- (89.82,209.82) -- cycle ;
    \draw  [fill={rgb, 255:red, 155; green, 155; blue, 155 }  ,fill opacity=1 ] (495.19,56.22) -- (512.82,56.22) -- (512.82,209.82) -- (495.19,209.82) -- cycle ;
    \draw    (109.82,234.64) -- (486.82,235.09) ;
    \draw [shift={(489.82,235.09)}, rotate = 180.07] [fill={rgb, 255:red, 0; green, 0; blue, 0 }  ][line width=0.08]  [draw opacity=0] (8.93,-4.29) -- (0,0) -- (8.93,4.29) -- cycle    ;
    \draw [line width=0.75]  [dash pattern={on 0.84pt off 2.51pt}]  (109.82,28) -- (190.82,28) ;
    \draw [line width=0.75]  [dash pattern={on 0.84pt off 2.51pt}]  (246.82,28) -- (322.82,28) ;
    \draw [line width=0.75]  [dash pattern={on 0.84pt off 2.51pt}]  (418.82,28) -- (458.82,28) ;
    
    \draw (138,119) node [anchor=north west][inner sep=0.75pt]  [color={rgb, 255:red, 255; green, 255; blue, 255 }  ,opacity=1 ] [align=left] {{\footnotesize \textbf{Encoder}}};
    \draw (272,119) node [anchor=north west][inner sep=0.75pt]  [color={rgb, 255:red, 255; green, 255; blue, 255 }  ,opacity=1 ] [align=left] {{\footnotesize \textbf{Transition}}};
    \draw (417,118) node [anchor=north west][inner sep=0.75pt]  [color={rgb, 255:red, 255; green, 255; blue, 255 }  ,opacity=1 ] [align=left] {{\footnotesize \textbf{Decoder}}};
    \draw (90,230) node [anchor=north west][inner sep=0.75pt]    {$\tilde{o}_{t}$};
    \draw (495,230) node [anchor=north west][inner sep=0.75pt]    {$\tilde{o}_{t+1}$};
    \draw (90,18) node [anchor=north west][inner sep=0.75pt]    {$\tilde{o}_{t}$};
    \draw (326,16) node [anchor=north west][inner sep=0.75pt]    {$P_{\theta _{s}} (s_{t+1} |\tilde{s}_{t} ,\tilde{a}_{t} )$};
    \draw (461,16) node [anchor=north west][inner sep=0.75pt]    {$P_{\theta _{o}} (o_{t+1} |\tilde{s}_{t+1} )$};
    \draw (196,16) node [anchor=north west][inner sep=0.75pt]    {$Q_{\phi _{s}}( s_{t})$};
    \draw (275,240) node [anchor=north west][inner sep=0.75pt]    {$\tilde{a}_{t} \ ,\ \Delta _{t} \ $};
    \end{tikzpicture}
    \caption{The agent's architecture and generative framework resemble that of a VAE. The line on top represents the agent simulating the future and making a prediction, while on the bottom, the agent receives a new observation after $\Delta _{t}$ of taking an action, $\tilde{a}_{t}$ .}
    \label{fig:agent}
\end{figure}
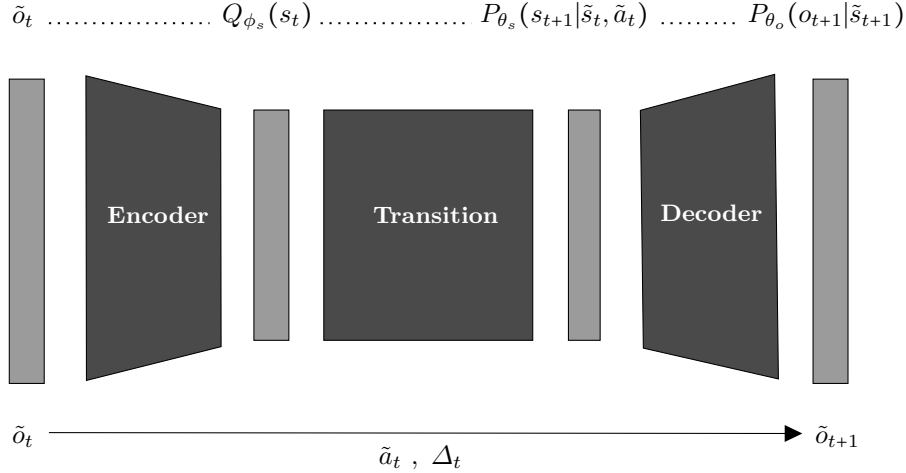

Utilizing the mentioned architecture (also depicted in Fig. \ref{fig:agent}), the agent would be able to calculate the EFE in Eq. \ref{eqn:efe} for a given policy (i.e., $\pi$), which is a sequence of actions. Thus, we can calculate the probabilities for selecting actions through Eq. \ref{eqn:2}. In order to make better decisions, the agent can plan ahead by simulating future trajectories using the architecture mentioned above. However, as the policy space grows exponentially into the future, it is infeasible to evaluate all possible scenarios. Fountas et al. (2020) \cite{Fountas2020DeepAI} introduced two approaches to alleviate the obstacle. 1) They used the standard Monte-Carlo Tree Search (MCTS) \cite{coulom2006efficient,silver2017mastering}, a tree-based search algorithm, which selectively explore promising trajectories in a restricted manner. 2) They introduced another recognition module \cite{piche2018probabilistic,marino2018iterative,tschantz2020control}, parameterized with ${\phi}_{a}$ as follows: \textbf{Habit} (i.e., $Q_{\phi_{a}}(a_{t})$), an amortized inference of actions (i.e., an inference network \cite{margossian2023amortized} providing a mapping between a sampled hidden state, $\tilde{s}_{t}$, and normalized, through \emph{Softmax} function, probabilities for the actions). 
Habit is also realized through a neural network, which approximates the posterior distribution over actions (i.e., ${P}({a}_{t}|{s}_{t})$) using the prior ${P}({a}_{t})$ that is obtained from the MCTS \cite{Fountas2020DeepAI}. This network is trained to reproduce the last action sampled form the planner, given the last state. This is similar to the fast and habitual decision-making in biological agents \cite{van2012information}. 

Fountas et al. (2020) \cite{Fountas2020DeepAI} followed the standard four steps of MCTS \cite{silver2017mastering,swiechowski2023monte}, which let them to restrict and prioritize the trajectories that should be evaluated. Iterativly a weighted tree that has memory updates the visited states. During each loop, a path from the existing tree (towards a leaf node) is selected (i.e., \textbf{selection}) based on the following upper bound confidence:
\begin{equation}
\label{eqn:mcts-ubc}
U(s_{t},a_{t})=\tilde{G}(s_{t},a_{t})+c_{\mathrm{explore}}\cdot Q_{\phi_{a}}(a_{t}|s_{t})\cdot\frac{1}{1+N(a_{t},s_{t})} \ .
\end{equation}
Where $\tilde{G}(s_{t},a_{t})$ represents the algorithm's current estimation for the EFE, while $N(a_{t},s_{t})$ denotes the number of times a node (i.e., a state and action pair) in the tree has been visited during the tenure, along with an exploration hyperparameter, $c_{\mathrm{explore}}$.
Then, starting from the leaf and for every possible action (i.e., \textbf{expansion}), the EFE is calculated for a fixed number of future steps (i.e., \textbf{simulation}). This value is finally added to all nodes along the path to calculate the average of $\tilde{G}(s_{t},a_{t})$ (i.e., \textbf{backpropagation}). After all, Actions are sampled from the probabilities created by $P(a_{t}) = \frac{N(a_{t},s_{t})}{\sum_{j}N(a_{t},j,s_{t})}$ (where $P(a_{t}) = \sum_{\pi:a_{1}=a_{t}}^{}P(\pi)$), which is proportional to the number of times a node is visited. The planning process continues until a maximum number of loops (i.e., a hyperparameter) is reached, or a condition, i.e., $\text{max} P(a_{t}) - \text{mean} P(a_{t}) > T_{dec}$, is met, indicating that the planning is finished.
Fountas et al. (2020) \cite{Fountas2020DeepAI} further employed $Q_{\phi_{a}}(a_{t})$ (i.e., habit) to modulate the state space, motivated by incorporating uncertainty. The divergence between the policy obtained from the planner (i.e., MCTS) and the habit (i.e., $D_{t}=D_{\mathrm{KL}}\left[Q_{\phi_{a}}(a_{t})\left|\right|P(a_{t})\right]$) can serve as a loss function for the habit. This divergence also represents a type of uncertainty in the state space, preventing the habit from being identical to the policy \cite{friston2017active,byers2012exploring}. Therefore, they used $D_{t}$ in a logistic function as follows:
\begin{equation}
\label{eqn:omega}
\bm{\omega}_{t}=\,\frac{\alpha}{1\,+\,e^{-\,\frac{b-D_{t-1}}{c}}}\,+\,d \ .
\end{equation} 
The monotonically decreasing pattern establishes a reciprocal connection between $D_{t-1}$ and $\bm{\omega}_{t}$, i.e., the state precision, using hyperparameters $\{\alpha, b, c, d\}$. Altogether, the transition (i.e., ${P}_{\theta_{s}}({s}_{t}|{s}_{t-1},{a}_{t-1})$) is modeled with the distribution $\mathcal{N}(\mu,\,\sigma^{2}/\bm{\omega}_{t})$ \cite{Fountas2020DeepAI}. This precision is analogous to $\beta$ in $\beta \text{-VAE}$ \cite{higgins2016beta}, effectively promoting disentanglement of the state space by the encoder \cite{Fountas2020DeepAI}. 

To facilitate the computations and effectively estimating different elements within the framework, Fountas et al. (2020) \cite{Fountas2020DeepAI} used several levels of Monte-Carlo simulations (i.e., sampling based estimations). In addition to MCTS, all terms in the EFE (Eq. \ref{eqn:efe}), including the use of MC dropout \cite{gal2016dropout} for model parameters (i.e., \ref{eqn:efe-c}), are estimated in this manner \cite{Fountas2020DeepAI}. 

\subsection{Enhancements}
\label{subsec:enhancements}

We explore various aspects to design the agent and leverage the formalism and existing architecture outlined by Fountas et al. (2020) \cite{Fountas2020DeepAI} as a foundation. First, we examine how features and requirements related to the problem under study can influence the agent. Subsequently, we propose solutions to address these issues, introducing a coherent agent design.

\subsubsection{Exploring the Application}
\label{subsubsec:exploring.application}
The simulation of the system described in Section \ref{subsec:eec}, is designed to replicate features of an industrial system. It utilizes Poisson processes \cite{kingman1992poisson} (i.e., exponential distributions) for machine state transitions and part arrivals \cite{LOFFREDO202391}. The \emph{event-driven} steps employed by Loffredo et al. (2023) \cite{LOFFREDO202391,loffredo2023reinforcement} trigger decisions after a system state transition rather than at fixed intervals, proving effective for controlling working machines. This problem can be viewed as either continuous-time stochastic control or a discrete-time Markov Chain process \cite{ross2014introduction}. Continuous-time modeling requires making time intervals visible to the agent for both machines and subsequent observations, whereas discrete-time modeling allows the agent to learn dynamics by observing transitions to create probabilities for different transitions. There is a variable time interval between subsequent observations (i.e., $\Delta t$ as depicted in Fig. \ref{fig:agent}). This variability requires synchronizing the transition for prediction (i.e., $P_{\theta _{o}} (o_{t+1} |\tilde{s}_{t+1} )$) with the next observation (i.e., $\tilde{o}_{t+1}$) in the continuous-time model. Incorporating continuous-time facilitates neural network function approximation for longer $\Delta t$ during planning. However, using continuous-time can further complicate the prediction structure since residence times for machine states exist in observation. Here, we utilize discrete-time \emph{event-driven} steps that simplify this process compared to the continuous-time approach.

The stochastic nature, along with the integral and continuous form of the reward functions for the system under study \cite{LOFFREDO202391,loffredo2023reinforcement}, implies that the effects of decisions may not be immediately observable, aligning with POMDPs.
The system has a delay in responding to the policy (i.e., delayed policy response), which we refer to as \emph{long/delayed impact horizon}, particularly with respect to reward. This is an important distinction from environments evaluated by Fountas et al. (2020) \cite{Fountas2020DeepAI}. The agent proposed by them focuses on planning based on immediate next predictions. The problem at hand is continuous with no terminal state, and the span over which reward functions are integrated may encompass a few thousand steps. This complicates planning and highlights the need for less computational cost.

\subsubsection{Experience Replay}
The generative model encompasses the core of active inference \cite{friston2010action,parr2022active,Fountas2020DeepAI} and predictive coding \cite{millidge2022predictive}. Therefore, the performance of any active inference-based agent heavily relies on its accuracy. To improve model training, we introduce experience replay \cite{mnih2015human} using a memory that stores $(o_{t},a_{t},o_{t+1})$ at different steps. During training, we sample a batch of experiences from the memory and ensure the latest experience is also included. However, for all the batched experiences, we utilize $\bm{\omega}_{t}$ based on the latest experience.

\subsubsection{Hybrid Horizon} 
To address the limitations arising from the short horizon of the EFE, which relies on immediate next predictions, we propose augmenting the planner with an auxiliary term to account for longer horizons. Q-learning \cite{watkins1992q} and its enhanced variant, deep Q-learning \cite{mnih2015human}, can serve as model-free planners with longer horizons, leveraging rewards to update the Q-value for a state-action pair. The Q-value represents the expected return, considering long-term consequences even in one-step lookahead \cite{sutton2018reinforcement}. Mnih et al. (2015) \cite{mnih2015human} demonstrated the effectiveness of DQN in learning relatively long-term strategies in certain games. Loffredo et al. (2023) \cite{LOFFREDO202391} demonstrated that deep Q-learning can achieve near-optimal performance for the systems under study. Accordingly, we modify \(Q_{\phi_{a}}(a_{t})\) to represent amortized inference of actions, mapping observations \(\tilde{o}_{t}\) (or sampled predictions) to normalized action probabilities using a \emph{Softmax} function and training it with deep Q-learning updates based on rewards from experience replay. We introduce a hyperparameter \(\gamma\) to balance the contributions of long and short horizons. Thus, we arrive at a new formulation for the planner to incorporate longer horizons:
\begin{equation}
\label{eqn:h-planner}
P(a_{t}) = \gamma \cdot Q_{\phi_{a}}(a_{t}) + (1-\gamma) \cdot {\sigma} \left( -G(\pi) \right) \ .
\end{equation}
The resulting combination achieves a controlled balance between the EFE for the short horizon terms, which incorporates uncertainties, and the long horizon term. We further utilize the new \(Q_{\phi_{a}}(a_{t})\) to modulate the agent's state uncertainty based on Eq. \ref{eqn:omega}. In fact, \(D_{t}\) (i.e., \(D_{\mathrm{KL}}\left[Q_{\phi_{a}}(a_{t})\left|\right|P(a_{t})\right]\)) represents the discrepancy between the long horizon and the combined policy, reflecting a form of knowledge gap for the agent.

\subsubsection{Multi-Step Transition and Planning}
Given the stochastic nature and \emph{long impact horizon} of the system under study, a one-step transition (as depicted in Fig. \ref{fig:agent}) may not result in significant changes in observation and state, leading to indistinguishable EFE terms. Therefore, the model should learn transitions beyond one step and predict further into the future to distinguish the impact of different policies. We modify the transition module to allow multiple steps, controlled by a hyperparameter (e.g., $s = 90$), enabling multi-step transitions given the policy (i.e., sequence of actions). Representing the sequence of actions in a policy as a one-hot vector can be high-dimensional, so we utilize integer encodings as an approximation. This is feasible since the actions (or number of machines) are less categorical and can be considered rather continuous in this case. During planning, we utilize repeated actions in the transition for each action and calculate the EFE accordingly. This method assesses the impact of actions over a short period, using repeated action simulations. This approximation helps to distinguish different actions over a horizon based on the EFE. Thus, even a single multi-step transition can serve as a simple and computationally efficient planner.
For deeper simulations, it can be combined with MCTS of repeated transitions. Alternatively, it can be combined with a less expensive planner that starts with repeated transitions for each action, followed by simulating until a specific depth using the following policy:
\begin{equation}
\tilde{P}(\pi, a_{\tau}) = (1 - c_{\mathrm{explore}}) \cdot Q_{\phi_{a}}(a_{\tau}) + c_{\mathrm{explore}} \cdot \sigma \left( -\log P(o_{\tau}|\pi) \right) \ ,
\label{eqn:light-plan}
\end{equation}
to calculate the EFE of the final state or trajectory. After several loops, the accumulated EFE for each action can be used in Eq. \ref{eqn:h-planner}.

\section{Results}
\label{sec:results}
To demonstrate the potential of our methodology, we concentrate our experiments on controlling a real industrial workstation comprising six parallel-identical machines with an upstream capacity of 10, as outlined in \cite{LOFFREDO202391}. All the stochastic processes characterizing the workstation follow Poisson distributions, and thus are exponentially distributed with different expected values. We first introduce the preference function, then describe the setup and training, and finally present the numerical performance of our agent\footnote{The source code is available at \href{https://github.com/YavarYeganeh/AIF_Meeting_EEC}{https://github.com/YavarYeganeh/AIF{\_}Meeting{\_}EEC}.}.

\subsection{Preference Function}
Active inference involves an agent acting to achieve its preferred observation, similar to the concept of a setpoint in control theory \cite{friston2017active,millidge2022predictive}. Consequently, the agent possesses an internal function to quantify the proximity of the prediction to the preferred state. This function is important for the agent's performance. While it correlates with the reward function in reinforcement learning, it is based on a different philosophy: a control setpoint rather than the cumulative reward in the MDP framework of reinforcement learning \cite{sutton2018reinforcement}. Instead of the reward function used by Loffredo et al. (2023) \cite{loffredo2023reinforcement}, we propose a different preference function for the multi-objective optimization of the system under study. This function aligns with the concept of a \emph{long/delayed impact horizon} system for our agent to control, accounting for the average performance of the system over a fixed time span (\( t_{s} \)\footnote{In our implementation, we use the closest recorded timestamp from the system, respecting the fixed time span.}, e.g., 8 hours) leading up to the observation at (\( t \)). 
It includes terms for production, energy consumption, and a combined term as follows:
\begin{subequations}
\begin{align}
R_{\text{production}} &= \frac{T_{\text{current}}}{T_{\text{max}}} \quad
R_{\text{energy}} = 1 - \frac{E_{\text{avg}}}{E_{\text{max}}} \\
R &= \phi \cdot R_{\text{production}} + \left(1 - \phi \right) \cdot R_{\text{energy}} \ ,
\end{align}
\label{eqn:preference}
\end{subequations}
where \( \phi \) is the weighting coefficient balancing production and energy consumption. \( \phi \) is set close to 1 (0.97 in our implementation) to ensure that the agent does not significantly reduce production. \( T_{\text{current}} =  \frac{NP(t) - NP(t - t_{s})}{t_{s}} \) represents the throughput within the past \( t_{s} \) period, where \( NP(t) \) is the number of parts produced within this period. \( T_{\text{max}} \) is the maximum achievable throughput, occurring under the \emph{ALL ON} policy \cite{loffredo2023reinforcement}. \( E_{\text{avg}} =  \frac{C(t) - C(t - t_{s})}{t_{s}} \) represents the average energy consumption over the past \( t_{s} \) period, where \( C(t) \) is the total energy consumption within this period. \( E_{\text{max}} \) is the maximum theoretical consumption rate of the system, pertaining to all machines operating in their busy state.

\subsection{Agent Setup and Training}
\label{subsec:Training}
We adhere to the MC sampling methodology for calculating EFE, as well as most agent configurations, as outlined by Fountas et al. (2020) \cite{Fountas2020DeepAI}. Notably, we employ Bernoulli and Gaussian distributions to model prediction and state distributions, respectively. We introduced a modification by utilizing activation functions for the encoder and transition networks. The output of these two networks generates means and variances representing Gaussian distributions for the state. We applied the \emph{Tangent Hyperbolic} function for the means and the \emph{Sigmoid} function, multiplied by a factor, $\lambda_s$, between 1 and 2, for the variances. This enforces further regularization and stability for the state space to prevent unbounded values, which need to be fitted into a normal distribution. In contrast to \cite{Fountas2020DeepAI}, which does not use activations for the state, our implementation includes these activations, as we found them essential for learning effective policies to achieve high rewards. 

Our agent's observation comprises buffer levels and machine states, all one-hot encoded. Similarly, we incorporate the three reward terms in Eq. \ref{eqn:preference}, predicting only their means without pre-defining a distribution. The agent's preference (i.e., $P(o_{\tau}|\pi)$) corresponds to the prediction of the combined reward term for the system, which the agent seeks to maximize. Given the composition of binary and continuous components of the observation, the loss of the VAE's reconstruction is equally scaled aggregation of binary cross-entropy and mean square error. Additionally, due to the one-hot structure of the observation, we sample the predictions, excluding the reward terms, which are treated as means, to be fed into the decoder during the calculation of EFE. As we validate the performance of a control system for an industrial application, we use a single system during each training epoch, with an experience replay size of 200. These systems are initialized with a random policy for one day of simulation, after removing the profile from one day of warm-up simulation using the \emph{ALL ON} policy. 
We trained our agents on-policy, following a similar approach to the algorithm proposed by Fountas et al. (2020) \cite{Fountas2020DeepAI}, but with adjustments to accommodate the modifications we made, including multi-step transition, experience replay, and a hybrid planner.

\subsection{Agent Performance}
\label{subsec:perfomcance}
\begin{figure}[!h]
    \centering
    \includegraphics[width=0.55\textwidth]{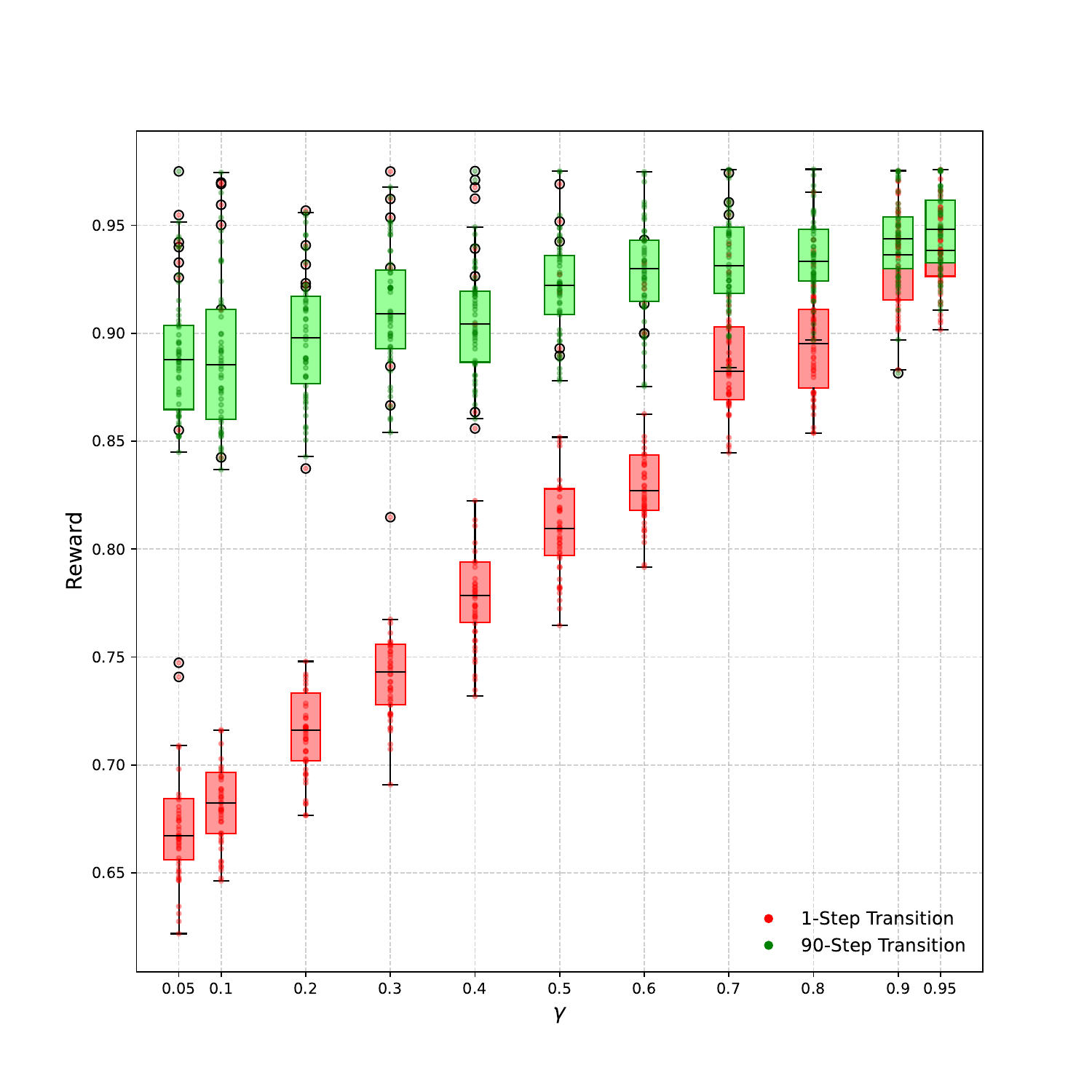}
    \caption{Comparison of test rewards during training of agents with 90-step transition against 1-step transition, when $\lambda_s=1$.}
    \label{fig:results1}
\end{figure}

To assess the efficacy of our agents, we tested their performance several times during different training epochs, each on independent systems initialized with a random agent after warm-up, similar to those used for training. We simulated the interaction between the agent and the system, which was randomly controlled at the beginning, for one day of simulation time. The combined reward term (in Eq. \ref{eqn:preference}) at the end of the test simulation was extracted as the performance measure, with a time span of \(t_s\) (i.e., 8 hours in this case). Our agent, similar to \cite{Fountas2020DeepAI}, has a large set of hyperparameters, which proved to be sensitive, especially those related to the state space of the model. We considered both 1-step transitions and multi-step transitions, taking repeated actions during planning for each of the possible actions (i.e., determining how many machines to keep \emph{ON}) to then calculate their EFE. Fig. \ref{fig:results1} presents the comparison for a single set of hyperparameters, particularly $\lambda_s=1$, except \(s\), across different $\gamma$. It shows the suitability of the multi-step transition and simple repeated planner as well as hybrid horizon.

\begin{figure}[!h]
    \centering
    \includegraphics[width=\textwidth]{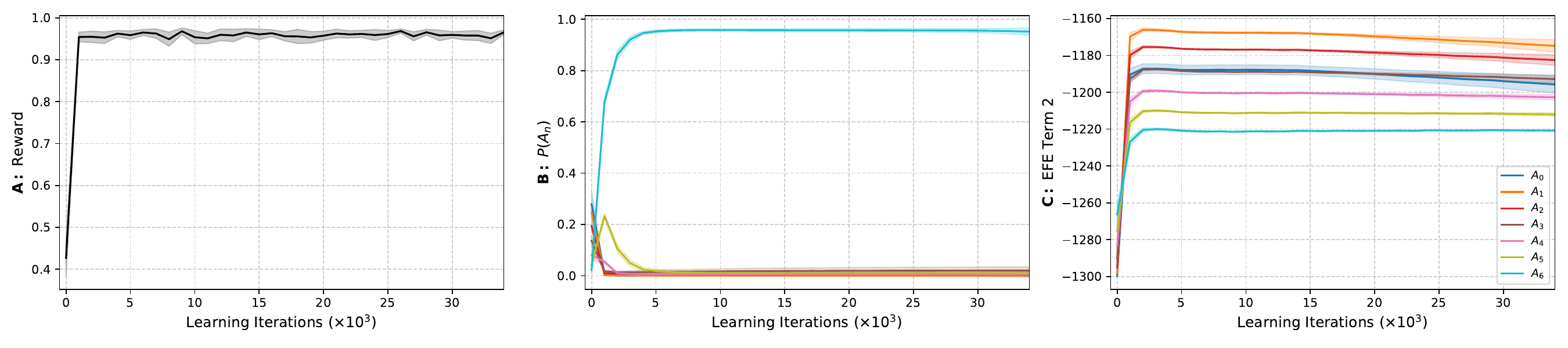}
    \caption{The test performance of the agent with 90-step transition and repeated actions for planning, when $\lambda_s=1.5$ and $\gamma=0.05$, replicated with 10 different random seeds. \textbf{A:} Average reward. \textbf{B:} Average planner distribution for different actions. \textbf{C:} Average EFE's term 2 for different actions.}
    \label{fig:results2}
\end{figure}

Since the state should fit the normal distribution, $\lambda_s=1$ for variance will target areas of the input domain of the \emph{Sigmoid} function that are saturated and have smaller gradients. To address this, we increased $\lambda_s$ to 1.5, where the \emph{Sigmoid} function has larger gradients. This results in higher rewards even for very small $\gamma$ (i.e., 0.05), as presented in Fig. \ref{fig:results2}. 
This performance is primarily coming from the second term of the EFE (i.e., Eq. \ref{eqn:efe-b}), which serves as an \emph{intrinsic reward} \cite{paul2023efficient}, as demonstrated in Fig. \ref{fig:results2}C. The other two terms are less distinguishable for different actions on average. This suggests that the agent differentiates between various actions in its state space to achieve high rewards. In fact, the \emph{long impact horizon} and stochasticity, as well as our approximations in transition and planning, hinder the predictive power of the agent, but it managed to infer the impact of different actions in its state space.
It is also worth noting that our agent quickly converges to high rewards but may experience instability and loss of control if training continues for (a long time), due to a \emph{catastrophic} increase in the loss, particularly the reconstruction term of the generative model. This necessitates early stopping mechanisms for training and the introduction of regularization elements \cite{goodfellow2016deep} (e.g., dropout and normalization) to prevent or mitigate the issue.

\section{Conclusion and Future Work}
The results demonstrate the effectiveness of our proposed modifications for the active-inference-inspired agent. Notably, a single multi-transition lookahead with repeated actions, coupled with a hybrid horizon, achieves high rewards without relying on extensive planning algorithms like MCTS. This is important given the application's need for deep lookaheads into the future, while stochasticity presents a challenge. 
Unlike RL, which generally depends on high-quality reward signals that can be expensive to collect and difficult to design, AIF operates directly on observations. This reduces the need for well-engineered and expensive reward structures while enabling more flexible and adaptive decision-making. Overall, the potential of our methodology for addressing the EEC problem and similar scenarios characterized by delayed policy response and high stochasticity is evident.

Improving the methodology could involve enhancing the generative model, the core of active inference, especially by introducing recurrent transitions or enhancing predictive capabilities. The integration of diffusion-based generative models instead of VAEs \cite{huang2024navigating} is also a promising direction. The framework and formalism of active inference agents show promise for non-stationary scenarios, where model-free agents may struggle to adapt swiftly.
Future research will focus on improving the agent, extending experimental validation, and tailoring the methodology for non-stationary scenarios, leveraging the strengths of active inference to develop more robust and efficient decision-making algorithms.

\section*{Acknowledgments}
The work presented in this paper was supported by HiCONNECTS, which has received funding from the Key Digital Technologies Joint Undertaking under grant agreement No. 101097296.

\bibliographystyle{splncs04}
\bibliography{references}
\end{document}